\documentclass{article}
\usepackage{url}
\usepackage{authblk}
\usepackage{amsmath}
\usepackage{amssymb}
\usepackage{float}
\usepackage{multirow}
\usepackage{array}
\usepackage{graphicx}
\usepackage{cite}
\usepackage{orcidlink}

\usepackage[numbers]{natbib}
\usepackage{hyperref}
\hypersetup{
    colorlinks=true, 
    pdfborder={0 0 0}, 
    citecolor=blue,
    linkcolor=blue,
    urlcolor=blue,
}

\title{Auto-US: An Ultrasound Video Diagnosis Agent Using Video Classification Framework and LLMs\vspace{1em}}

\author[1]{Yuezhe Yang \orcidlink{0009-0002-0059-0365} }
\author[1]{Yiyue Guo}
\author[1]{Wenjie Cai}
\author[2]{Qingqing Ruan}
\author[1]{Siying Wang}
\author[1,*]{Xingbo Dong \orcidlink{0000-0001-9782-6068} }
\author[1]{Zhe Jin}
\author[3,4]{Yong Dai \orcidlink{0000-0002-6840-9158} }
\affil[1]{The Anhui Provincial International Joint Research Center for Advanced Technology in Medical Imaging, Anhui University, Hefei, 230093, China.}
\affil[2]{ School of First Clinical Medicine, Anhui University of Science and Technology, Huainan, 232001, China.}
\affil[3]{School of Medicine, Anhui University of Science and Technology, Huainan, 232001, China}
\affil[4]{The First Hospital, Anhui University of Science and Technology, Huainan, 232001, China \vspace{1em}}
\affil[*]{Corresponding author. E-mail: \href{mailto:xingbo.dong@ahu.edu.cn}{xingbo.dong@ahu.edu.cn }; 
\authorcr
Contributing authors: \href{mailto:wa2214014@stu.ahu.edu.cn}{wa2214014@stu.ahu.edu.cn}; \href{mailto:wa2224013@stu.ahu.edu.cn}{wa2224013@stu.ahu.edu.cn}; \href{mailto:wa2214030@stu.ahu.edu.cn}{wa2214030@stu.ahu.edu.cn}; \href{mailto:17717324676@163.com}{17717324676@163.com}; \href{mailto:w124301145@stu.ahu.edu.cn}{w124301145@stu.ahu.edu.cn};
\href{mailto:jinzhe@ahu.edu.cn}{jinzhe@ahu.edu.cn};
\href{mailto:daiyong22@aust.edu.cn}{daiyong22@aust.edu.cn}}

\date{}

\begin{document}

\maketitle

\vspace{1em}

\begin{abstract}
AI-assisted ultrasound video diagnosis presents new opportunities to enhance the efficiency and accuracy of medical imaging analysis. However, existing research remains limited in terms of dataset diversity, diagnostic performance, and clinical applicability. In this study, we propose \textbf{Auto-US}, an intelligent diagnosis agent that integrates ultrasound video data with clinical diagnostic text. To support this, we constructed \textbf{CUV Dataset} of 495 ultrasound videos spanning five categories and three organs, aggregated from multiple open-access sources. We developed \textbf{CTU-Net}, which achieves state-of-the-art performance in ultrasound video classification, reaching an accuracy of 86.73\% Furthermore, by incorporating large language models, Auto-US is capable of generating clinically meaningful diagnostic suggestions. The final diagnostic scores for each case exceeded 3 out of 5 and were validated by professional clinicians. These results demonstrate the effectiveness and clinical potential of Auto-US in real-world ultrasound applications. Code and data are available at: \url{https://github.com/Bean-Young/Auto-US}.
\end{abstract}


\section{Introduction}
\label{sec1}

Ultrasound imaging is a powerful tool in medical diagnostics, offering numerous advantages such as safety, broad adaptability, repeatability, strong soft tissue contrast, and cost-effectiveness \cite{9399640}. Consequently, ultrasound imaging serves as a widely used diagnostic tool for assessing internal body structures, including abdominal organs, the cardiovascular system, the urinary system, and fetal development \cite{wells1999ultrasonic}. 

Given these capabilities, ultrasound imaging has become an indispensable technique in medical diagnosis \cite{seo2017ultrasound}. Physicians can leverage the images and videos generated by ultrasound examination, combined with the patient's symptoms and medical history, to precisely locate lesions, clarify their size, shape, and relationship with surrounding tissues \cite{LEVIN2004549}. This process provides a key basis for diagnosing a wide range of diseases.

Historically, ultrasound video diagnosis has largely relied on the experience of clinicians, who use common features in ultrasound images to diagnose different pathological conditions. Since its origin in the late 1940s, medical ultrasound technology has undergone key breakthroughs such as mechanical scanning, grayscale imaging, and color Doppler blood flow technology, becoming the second most widely used medical imaging method after X-ray \cite{shung2011diagnostic}. Various conditions present specific abnormalities in ultrasound imaging, such as pneumonia \cite{parlamento2009evaluation,volpicelli2021lung}, gallbladder disease \cite{gerstenmaier2016contrast}, and breast tumors \cite{kobayashi1985echographic,solivetti2019sonographic}.

However, there are some obvious challenges to medical ultrasound imaging diagnosis. Low image quality due to noise and artifacts, along with significant variability across ultrasound systems from different institutions and manufacturers, makes ultrasound interpretation highly dependent on the expertise of experienced physicians or diagnostic specialists \cite{LIU2019261}. What's more, these issues are particularly evident in areas with scarce medical resources \cite{HUANG2023126790}. Additionally, current ultrasound video diagnosis is highly influenced by the subjectivity of doctors, which can lead to inconsistent interpretations.

Fortunately, with technological advancements, artificial intelligence(AI)-assisted medical ultrasound imaging diagnosis has emerged as a promising solution. Initially, convolutional neural networks (CNNs) became a crucial tool in medical image analysis due to their powerful feature extraction capabilities \cite{LITJENS201760}. CNNs can automatically extract features from complex ultrasound images, assisting physicians in manual diagnosis \cite{YU202192}. 

Nevertheless, researchers have found that CNNs struggle to capture critical contextual information essential for accurate diagnosis when processing dynamic or continuous ultrasound videos \cite{zhang2019spatio}. In contrast, the Transformer \cite{vaswani2017attention} architecture has gained widespread attention due to its ability to effectively capture long-range dependencies and global information through the self-attention mechanism \cite{Ersavas2024}. Researchers have proposed numerous Transformer variants to further enhance their capabilities in computer vision \cite{lin2022survey}.

Among them, Vision Transformer (ViT) \cite{dosovitskiy2020image} has demonstrated significant advantages in medical ultrasound image diagnostics. ViT segments images into fixed-size patches and uses the self-attention mechanism of Transformer to capture global information. This makes it particularly useful for understanding complex backgrounds and identifying features that span larger regions. Despite their advantages, ViT also face certain limitations when applied to dynamic sequential data. Particularly, the lack of temporal modeling capabilities and the high computational cost of self-attention pose significant challenges to its application in ultrasound videos \cite{dosovitskiy2020image}.

To facilitate the effective application of deep learning in video classification, various models have been developed. One of the earliest approaches, Convolutional 3D Networks (C3D) \cite{ji20123d}, employs 3D convolutions to extract spatiotemporal features, capturing both spatial and temporal information for video understanding. To further enhance spatiotemporal representation, SlowFast \cite{feichtenhofer2019slowfast} introduces a dual-pathway architecture: a slow branch that captures rich temporal semantics, and a fast branch that models short-term dynamic changes, enabling effective feature fusion. Building on the Transformer framework, TimeSformer \cite{bertasius2021space} employs a decomposed self-attention mechanism to model long-range dependencies in both spatial and temporal dimensions directly. This approach eliminates the computational overhead of 3D convolution while preserving the ability to capture global information.

From a frequency perspective, researchers have found that ViT has poor performance in capturing high-frequency components \cite{bai2022improving}. This is because ViT divides the image into a series of image blocks and thus cannot effectively use the local structure, which is more related to high-frequency components. To address this limitation, researchers have proposed architectures like Spectrformer \cite{patro2023spectformer}, which integrates spectrum and multiple attention layers within a transformer framework, and frequency channel-attention based vision Transformer methods \cite{xiang2025frequency} that leverage frequency domain channel-attention mechanisms.

Thus, deep learning models have made significant progress in video classification tasks. However, when applied to ultrasound imaging diagnostics, \textbf{three major challenges remain}.

First, we observe that the application of deep learning in ultrasound video classification remains quite narrow. For example, Howard et al. \cite{howard2020improving} applied deep learning to echocardiography, exploring novel CNN architectures and demonstrating that time-distributed and two-stream networks significantly improve accuracy by leveraging motion information throughout the cardiac cycle. In pulmonary ultrasound imaging, Shea et al. \cite{shea2023deep}, Ebadi et al. \cite{ebadi2021automated}, and Barros et al. \cite{barros2021pulmonary} applied deep learning models for pneumonia diagnosis. Meanwhile, Zhao et al. \cite{zhao2023deep}, Zhang et al. \cite{zhang2024using}, and Sun et al. \cite{sun2023boosting} focused on breast ultrasound. These studies are restricted to specific organ-based ultrasound imaging for particular disease diagnoses, limiting their generalizability for broader clinical ultrasound applications.

Furthermore, while existing video classification models achieve remarkable performance improvements on large-scale natural image tasks, they often overlook the unique characteristics of ultrasound medical imaging \cite{karpathy2014large, yue2015beyond}. These models typically demand substantial computational and storage resources, which hinders their deployment in clinical applications \cite{yuan2024vit}. We argue that to better understand ultrasound videos and effectively perform classification and diagnostic tasks, it is essential to incorporate spatial, temporal, and frequency-domain information. In particular, the high-frequency components of ultrasound images can play a crucial role in identifying lesion boundaries.

Finally, data plays a critical role in deep learning. Since deep learning fundamentally relies on learning patterns and features from data, the quality, quantity, and diversity of the dataset directly impact the model's performance and generalization ability \cite{brattain2018machine}. If the training data is insufficient or lacks representativeness, the model may suffer from overfitting. However, existing ultrasound video datasets are often collected from specific regions or provided by individual hospitals \cite{Yang_2023_ICCV, zhang2024new}. Due to medical data privacy regulations, assembling a multi-center heterogeneous dataset is highly challenging \cite{yang2025annotated}. We believe that constructing a diverse, heterogeneous dataset is essential for advancing ultrasound video analysis.

Meanwhile, we have observed the rapid development of large language models (LLMs), which can significantly enhance medical-assisted diagnosis \cite{yang2025application}. These models enable interactive dialogue and have demonstrated near-human or even human-level performance across various cognitive tasks, including those in the medical field \cite{thirunavukarasu2023large}. For instance, recent studies by Boussina et al. \cite{boussina2024large}, Aali et al. \cite{aali2024dataset}, and Goswami et al. \cite{goswami2024parameter} have explored the integration of large language models into medical diagnosis, demonstrating that their incorporation into reporting systems and clinical decision support can significantly enhance the quality and efficiency of medical services.

In the context of medical diagnosis, LLMs have shown promising results. Zhang et al. used Kaggle medical data to fine-tune GPT-2 models and develop systems in conjunction with medical databases \cite{zhang2024chatbot}, achieving 97.50\% accuracy and a 99.91\% AUC value in interpreting symptoms and providing diagnostic suggestions. Another study showed GPT-4 was comparable to radiologists in detecting errors in radiology reports and had the potential to reduce processing time and costs \cite{gertz2024potential}.

Specifically for ultrasound diagnosis, one study analyzed data from three hospitals in China, including 400 ultrasound reports and 243 errors \cite{yan2025use}. The findings indicate that LLMs have significant potential to enhance the accuracy of ultrasound reporting and, in some cases, even outperform human experts. Recently, DeepSeek \cite{guo2025deepseek}, an open-source LLM, has demonstrated strong problem-solving abilities through chain-of-thought reasoning \cite{peng2025gpt}. Although DeepSeek faces challenges with visual tasks \cite{jegham2025visual}, when users describe visual features of a disease in text, it can combine this information with other data for reasoning \cite{diniz2025deepseek}, making it potentially valuable for ultrasound diagnosis when combined with video classification models.

This inspires us to explore the possibility of combining deep learning networks with LLMs to assist in ultrasound image diagnosis.

Based on the above analysis, we propose \textbf{Auto-US}, the first agent that combines deep learning models with LLMs for ultrasound video-assisted diagnosis. Our contributions include:

\begin{enumerate}
    \item We are the first to attempt the construction of a multi-modal ultrasound diagnostic agent that integrates ultrasound videos with clinical texts.
    \item We designed a novel CNN-Transformer network that fully analyzes the temporal, spatial, and frequency-domain information in ultrasound videos.
    \item We constructed a comprehensive dataset that integrates multiple existing open access data sources, covering five typical disease categories and three different body parts.
    \item Compared with other video classification networks, our model achieved state-of-the-art (SOTA) performance, with an accuracy of 86.73\%.
    \item By integrating large language models, we conducted case studies in clinical settings, demonstrating the feasibility for assisted diagnosis.
\end{enumerate}

The structure of the paper is as follows: Section \ref{sec4} presents our experimental settings and results in detail; Section \ref{sec5} offers an in-depth discussion of the findings and outlines the limitations; and Section \ref{sec3} provides a detailed description of the methodology for constructing the Auto-US agent.

\section{Result}
\label{sec4}

\subsection{Datasets}

We constructed a comprehensive ultrasound video dataset (\textbf{CUV Dataset}) by integrating multiple public datasets. This integration resulted in a dataset comprising 495 ultrasound videos, which were carefully categorized into five distinct classes. The diverse nature of this integrated dataset allowed us to train and evaluate our model on a wide range of medical conditions. At the same time, to assess the clinical feasibility of Auto-US, we employed two clinical cases for the Case Study, evaluating the agent’s overall capability.

\subsection{CUV Dataset}

Based on the standardized processing workflow described in Section \ref{sec3.1}, CUV Dataset was constructed by integrating four publicly available ultrasound video datasets: \textit{BUV Dataset} \cite{10.1007/978-3-031-16437-8_59}, \textit{UIdataGB} \cite{turki2024uidatagb}, \textit{POCUS} \cite{born2020pocovid}, and \textit{Butterfly} \cite{chen2021uscl}.

The resulting comprehensive dataset encompasses five disease categories spanning three distinct organs: \textbf{benign breast lesions}, \textbf{malignant breast lesions}, \textbf{gallbladder diseases}, \textbf{COVID-19 pneumonia}, and \textbf{bacterial pneumonia}. This design ensures sufficient pathological and anatomical diversity for developing and benchmarking robust multi-disease ultrasound video classification models. 

\begin{figure}[htbp]
    \centering
    \includegraphics[width=\textwidth]{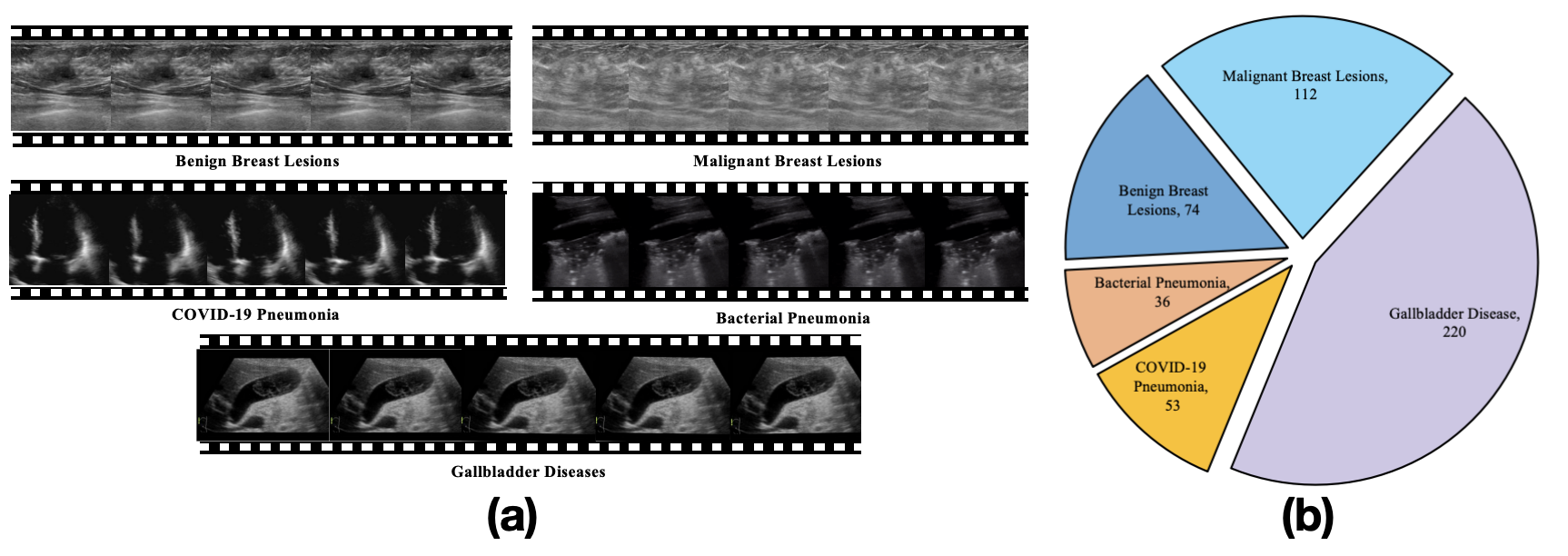} 
    \caption{Structure and Composition of the CUV Dataset. (a) Overview of sample categories in the CUV Dataset. (b) Pie chart of data distribution in the CUV Dataset.}
    \label{fig1}
\end{figure}

The labels and sample counts contributed by each source dataset toward the construction of our final comprehensive dataset are summarized in Table \ref{tab:1}. 
And the structure and data distribution of our comprehensive can be seen in Figure~\ref{fig1}. Specifically, Figure~\ref{fig1}\textbf{(a)} provides an overview of sample categories in the comprehensive dataset, while Figure~\ref{fig1}\textbf{(b)} shows a pie chart of data distribution in the comprehensive dataset.

\begin{table}[H]
\centering
\caption{Summary of source datasets and their contributions.}
\label{tab:1}
\begin{tabular}{
>{\centering\arraybackslash}p{4cm}|
{c}|{c}
}
\hline\hline
\multirow{1}{*}{Dataset} 
& \multicolumn{1}{c|}{Class} & \multicolumn{1}{c}{Sample}  \\
\hline
BUV Dataset  & Benign / Malignant Breast Lesions & 74 / 112 \\
UIdataGB  & Gallbladder Disease & 220 \\
POCUS & COVID-19 / Bacterial Pneumonia & 33 / 36 \\
Butterfly & COVID-19 Pneumonia & 20 \\
\hline\hline
\end{tabular}
\end{table}

\subsection{Experiment Setup}

For the CUV Dataset, 80\% were allocated for training purposes, while the remaining 20\% were reserved for testing. Each video was subjected to a classification model, which generated a single label from the categories of benign breast lesions (Benign), malignant breast lesions (Malignant), gallbladder disease (Gall.), COVID-19 (COVID), and bacterial pneumonia (Pneu.).

All experiments were conducted on 4 Nvidia RTX 3090 GPU with 24GB of RAM, using Python 3.8 and PyTorch 2.0. The input image size was set to 224$\times$224, and the batchsize on each GPU was set to 4. During training, the patch size is 4, and we used a widely-used Adam \cite{kingma2014adam} optimizer with a momentum of 0.9 and weight decay of 1e-4 for model optimization during backpropagation.

\subsection{Evaluation Metrics}

We evaluate the performance of our model using four standard metrics: Accuracy, Precision, Recall, and the Area Under the ROC Curve (AUC) for each class. Macro-averaged scores are adopted for multi-class evaluation, ensuring that each class is given equal weight regardless of its frequency.

\subsection{Quantitative Evaluation}

We designed a CNN-Transformer architecture tailored for Ultrasound video classification, named \textbf{CTU-Net}.

\begin{figure}[H]
    \centering
    \includegraphics[width=\textwidth]{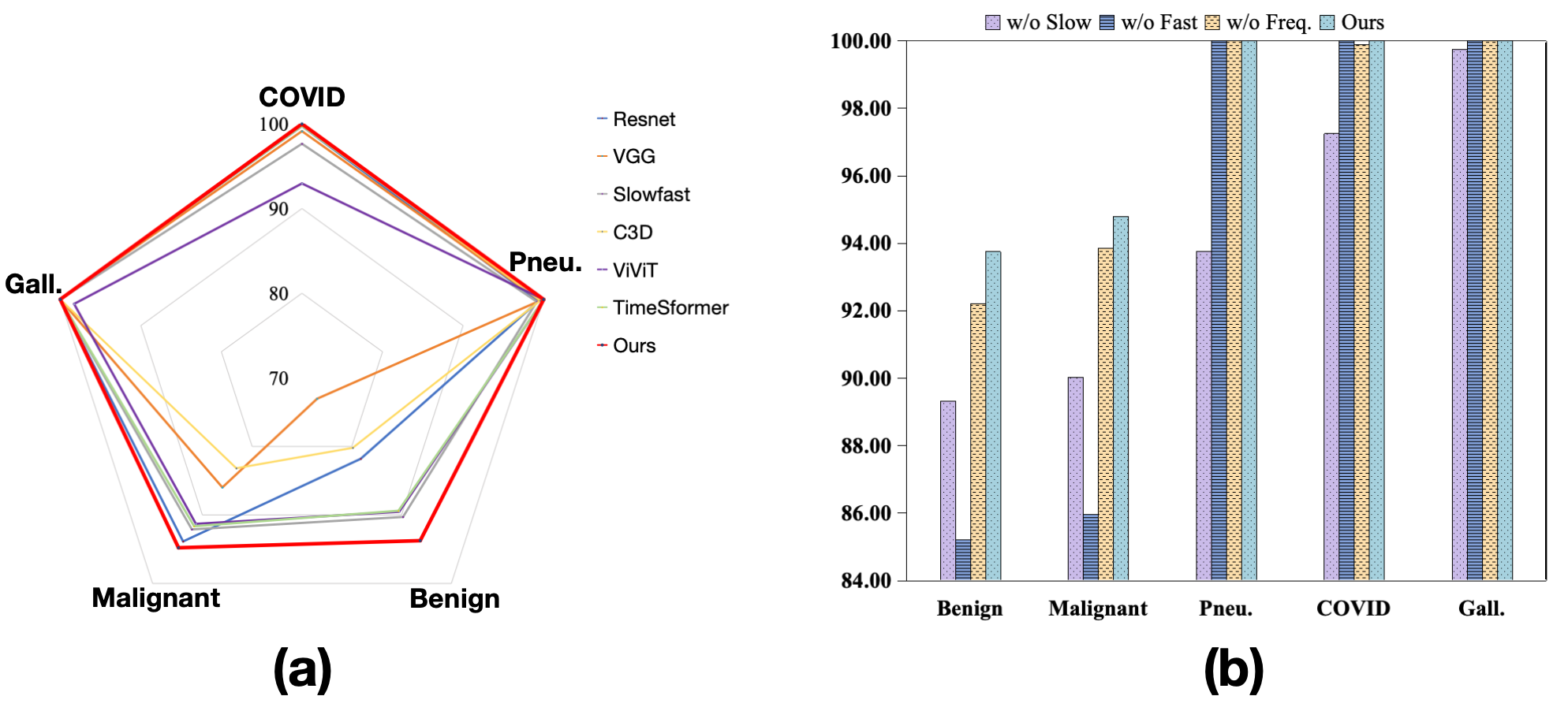} 
    \caption{AUC Distributions and Ablation Study of CTU-Net. \textbf{(a)} Radar chart of AUC across categories for different models. \textbf{(b)} Category-wise AUC comparison from CTU-Net ablation experiments.}
    \label{fig3}
\end{figure}

We evaluate the performance of CTU-Net by comparing it with several representative baselines, including CNN-based models ResNet \cite{tran2018closer} and VGG \cite{simonyan2014two}, temporal modeling approaches SlowFast \cite{Feichtenhofer_2019_ICCV} and C3D \cite{ji20123d}, as well as Transformer-based methods including ViViT \cite{arnab2021vivit} and TimeSformer \cite{bertasius2021space}. The results are summarized in Table~\ref{tab:2} and illustrated in Figure~\ref{fig3}\textbf{(a)}. It demonstrates the superior performance of CTU-Net across multiple categories and key evaluation metrics. Specifically, CTU-Net outperforms existing approaches in terms of accuracy, recall, and precision, achieving an accuracy of 86.73\%, which is notably higher than the next best model, TimeSformer. Furthermore, it achieves a 25.62\% improvement in recall compared to SlowFast and a 10.08\% increase in precision over VGG.

Figure~\ref{fig3}\textbf{(a)} further supports these improvements, demonstrating that CTU-Net consistently achieves high performance across all categories. Notably, it attains the highest AUC values among the compared methods, highlighting its effectiveness in accurately classifying and identifying relevant medical conditions in ultrasound videos. This capability is particularly important in clinical settings, where accurate and timely diagnosis is essential.

\begin{table}[H]
\centering
\caption{Comparison of performance metrics for different models. Bold is the best.}
\label{tab:2}
\begin{tabular}{
>{\centering\arraybackslash}p{3cm}|
*{3}{c}
}
\hline\hline
\multirow{1}{*}{\textbf{Method}} 
& \multicolumn{1}{c}{Accuracy(\%)} & \multicolumn{1}{c}{Recall(\%)} & \multicolumn{1}{c}{Precision(\%)}  \\
\hline
Resnet & 82.83 & 78.14 & 82.16 \\
VGG  & 80.80 & 77.94 & 75.39 \\
\hline
\rule{0pt}{4mm}
Slowfast & 77.78 & 59.13 & 78.37 \\
C3D & 79.80 & 78.42 & 75.56 \\
\hline
\rule{0pt}{4mm}
ViVit & 83.84 & 80.34 & 81.27 \\
TimeSformer  & 84.85 & 84.67 & 83.11 \\
\hline
\rule{0pt}{4mm}
CTU-Net (Ours) & \textbf{86.73} & \textbf{84.75} & \textbf{85.47} \\
\hline\hline
\end{tabular}
\label{tab:2}
\end{table}

\subsection{Ablation Study}

In this section, we conducted an ablation study to assess the contribution of each path in CTU-Net to its overall performance in ultrasound video diagnosis. The results, presented in Table~\ref{tab:3} and Figure~\ref{fig3}\textbf{(b)}, highlight the impact of removing the slow path, fast path, and frequency path on key metrics.

The experimental results indicate that removing the slow path led to a significant decrease in performance, with recall dropping to 56.45\% and precision to 51.99\%. This underscores the importance of the slow path, which captures essential spatial features, in maintaining high diagnostic accuracy in ultrasound video analysis.

Omitting the fast path resulted in performance degradation, with accuracy dropping by 3.06\%. This demonstrates that capturing temporal information in ultrasound imaging is critical for achieving high-precision ultrasound video classification.

Finally, removing the frequency path resulted in a 5.83\% decrease in recall, indicating that simple fusion of temporal and spatial domains is insufficient for precise feature extraction. This limitation becomes especially evident when dealing with class imbalance in ultrasound imaging, leading to less accurate classification results.

An analysis of the AUC values for each category further demonstrates that CTU-Net achieves the best performance across all categories. This reinforces the fact that all three paths are crucial for achieving precise ultrasound video classification. These findings highlight that a model that integrates spatial, temporal, and frequency domain information is essential for accurate feature extraction. And it forms a critical foundation for ultrasound diagnosis.

\begin{table}[H]
\centering
\caption{Ablation analysis of different paths of CTU-Net.}
\begin{tabular}{
>{\centering\arraybackslash}p{3cm}|
*{3}{c}
}
\hline\hline
\multirow{1}{*}{\textbf{Method}} 
& \multicolumn{1}{c}{Accuracy(\%)} & \multicolumn{1}{c}{Recall(\%)} & \multicolumn{1}{c}{Precision(\%)}  \\
\hline
w/o Slow & 75.51 & 56.45 & 51.99 \\
w/o Fast  & 83.67 & 81.91 & 82.03 \\
w/o Freq. & 85.86 & 78.92 & 83.73 \\
Full CTU-Net & \textbf{86.73} & \textbf{84.75} & \textbf{85.47} \\
\hline\hline
\end{tabular}
\label{tab:3}
\end{table}


\subsection{Case Study}

\begin{table}[t]
\centering
\scriptsize
\caption{Clinical Case 1 and Model Output Analysis}
\begin{tabular}{
>{\raggedleft\arraybackslash}p{1cm}
>{\arraybackslash}p{10cm}             
}
\hline\hline
\textbf{Input:} &\textbf{Ultrasound Imaging 
Diagnosis Opinion:} Malignant breast lesion 

\textbf{Chief Complaint:} A mass in the left breast was discovered for 1 week.

\textbf{Physical Examination:} Both breasts are symmetrical, with the two nipples at the same height. A mass approximately 1*2cm in size can be seen in the outer upper quadrant of the left breast. It is medium in texture, with unclear boundaries, good mobility, no local skin redness or swelling, no local skin ulceration, no nipple discharge, no orange peel-like changes, no nipple depression, and no lymph node enlargement is palpated under the armpit.

\textbf{Additional Information:} A 48-year-old female. \\
\hline
\textbf{Output:} &\textbf{Preliminary Diagnosis:} The findings are highly suggestive of invasive cancer (BI-RADS class 4B). Other possibilities such as fibroadenoma or phyllodes tumor are less likely due to malignant ultrasound features.

\textbf{Justification:} A mass of 1×2cm in the upper outer quadrant of the left breast with unclear boundaries, medium texture, and good mobility was found on ultrasound (BI-RADS class 4B). The patient is 48 years old with no nipple discharge, local skin redness, or palpable lymph node enlargement in both armpits. Pathological confirmation (e.g., biopsy) is urgent as mobile masses may indicate early malignancy.

\textbf{Recommended Follow-Up Examinations:} 

1.Complete preoperative evaluation, including routine blood tests (three major routines: CBC, blood biochemistry, coagulation profile), tumor marker assessment (e.g., CA15-3, CEA), and chest CT to rule out pulmonary metastases or pleural abnormalities.

2.Multidisciplinary tumor board review involving senior physicians (oncology, breast surgery, radiology) to confirm the diagnosis and finalize the surgical/adjuvant treatment plan.

3.Prompt surgical intervention (e.g., lumpectomy or mastectomy with sentinel lymph node biopsy) if no contraindications (e.g., metastatic disease, severe comorbidities). Postoperative adjuvant therapy (endocrine therapy, chemotherapy) should be planned based on final pathology and receptor status (ER/PR-positive, HER2-negative).
\\
\hline
\textbf{Grades:} & 
Amateur I: 4/5,\quad Amateur II: 5/5,\quad Amateur III: 2/5,\\
& Physician I: 4/5,\quad Physician II: 3/5, \quad METEOR: 0.42 \\
\hline

\textbf{Analysis:} & Final Score: 3.25/5 

In this case, the Auto-US model accurately identified key ultrasound features of a potentially malignant breast mass and provided a reasonable preliminary diagnosis of invasive cancer, along with appropriate recommendations such as biopsy and preoperative evaluation, demonstrating strong early diagnostic and decision-support capabilities. However, it lacked critical details found in the GroundTruth, including pathology and immunohistochemistry results (e.g., ER, PR, HER2, Ki-67) that are essential for confirming cancer type and guiding treatment. It also did not consider alternative diagnoses like intraductal papilloma or fibroadenoma. While effective for initial assessment, the model requires enhancement in handling complex cases, particularly through integration of pathology and molecular data. \\

\hline\hline
\end{tabular}
\label{tab:4}
\end{table}

\begin{table}[t]
\centering
\scriptsize
\caption{Clinical Case 2 and Model Output Analysis}
\begin{tabular}{
>{\raggedleft\arraybackslash}p{1cm}
>{\arraybackslash}p{10cm}             
}
\hline\hline
\textbf{Input:} &\textbf{Ultrasound Imaging 
Diagnosis Opinion:} Gallbladder disease

\textbf{Chief Complaint:} A fever of 3 days.

\textbf{Physical Examination:} After hospitalization, physical examination revealed tenderness in the abdomen, and a history of gallbladder stones was found during a physical examination. 

\textbf{Additional Information:} An elderly woman has many underlying diseases and urinary tract infections. \\
\hline
\textbf{Output:} &\textbf{Preliminary Diagnosis:} The patient presents with acute cholecystitis complicated by suppurative transformation, likely evolving from preexisting cholelithiasis. Clinical evidence suggests possible progression to empyema or necrotizing changes. Secondary diagnostic considerations should include ascending cholangitis and sepsis syndrome, given concurrent urinary tract infection and underlying comorbidities including hypertension and diabetes mellitus.

\textbf{Justification:} The diagnosis of acute cholecystitis with cholelithiasis is established through characteristic clinical triad (fever, right upper quadrant tenderness, leukocytosis) consistent with Tokyo Guidelines criteria. The patient's advanced age and comorbid conditions predispose to atypical disease manifestations and higher risks of gallbladder necrosis or systemic complications. Differential diagnosis remains broad pending exclusion of biliary tree involvement and confirmation of infection focus.

\textbf{Recommended Follow-Up Examinations:} 
To ensure appropriate management, essential investigations include serial inflammatory markers (CBC, CRP, procalcitonin), hepatic panel, and blood cultures to assess the extent of infection. Cross-sectional imaging with contrast-enhanced abdominal CT is prioritized for evaluating gallbladder integrity and detecting potential complications. MRCP should be obtained if biochemical evidence of cholestasis develops. Urine culture and renal function tests are required to characterize urinary tract involvement. Immediate initiation of empiric antimicrobial therapy coupled with surgical consultation for definitive biliary source control is warranted, along with optimization of glycemic control and hemodynamic parameters. \\
\hline
\textbf{Grades:} & 
Amateur I: 4/5,\quad Amateur II: 5/5,\quad Amateur III: 3/5,\\
& Physician I: 4/5,\quad Physician II: 3/5, \quad METEOR: 0.39 \\
\hline

\textbf{Analysis:} & Final Score: 3.29/5 

The Auto-US model provides a structured diagnostic framework by integrating clinical signs and patient comorbidities. Its preliminary diagnosis of acute cholecystitis with suppurative transformation is consistent with typical presentations in elderly patients with gallstones. It also appropriately considers differential diagnoses such as ascending cholangitis and sepsis, reflecting the complexity of systemic infections in high-risk populations. However, it omits key clinical evidence from the GroundTruth, such as purulent drainage findings, specific lab results, and final discharge diagnoses including pyelonephritis, hypertension, and diabetes, making it less comprehensive in clinical detail. \\
\hline\hline
\end{tabular}
\label{tab:5}
\end{table}
We selected two distinct clinical case studies, involving a malignant breast tumor and gallbladder disease, with detailed results shown in Table \ref{tab:4} and Table \ref{tab:5}. These cases demonstrate that the Auto-US model exhibits significant advantages in clinical decision support, particularly in generating preliminary diagnoses and treatment recommendations based on ultrasound video analysis and relevant clinical context.

We observed considerable variation in the evaluations of Auto-US outputs across different individuals. Some amateur evaluators considered the model a valuable assistant for guiding diagnosis and clinical reasoning, even suggesting it could potentially replace physicians in some scenarios. However, others noted only partial keyword overlap. Among professional physicians, evaluations ranged from “marginally acceptable” to “useful for diagnostic assistance,” highlighting the model’s promising potential in real-world clinical applications.

Despite this, the METEOR scores remained relatively low, likely due to the model’s strict handling of synonyms and occasional omission of fine details. The complexity of medical contexts inherently challenges the achievement of high METEOR scores.

We also found that Auto-US still lacks comprehensive coverage of complex diagnostic pathways, particularly in cases requiring pathological, molecular testing, or fine-grained differential diagnosis. Future development should focus on the integration of multi-modal models, incorporating pathology and other modalities to provide more accurate and complete diagnostic recommendations.

\subsection{Conclusion}
\label{sec6}

In this work, we introduced \textbf{Auto-US}, the first ultrasound video diagnostic agent. Our approach includes the construction of a comprehensive dataset. We also propose a novel deep learning network for ultrasound video classification. This network fully leverages temporal, spatial, and frequency-domain information. Additionally, we integrate LLMs to enhance the diagnostic process. This integration highlights the potential of Auto-US to deliver accurate and efficient diagnostic suggestions, offering valuable insights for clinical decision-making. We believe that Auto-US can serve as an important reference for future research in the field of automated ultrasound diagnosis, paving the way for more advanced and reliable diagnostic tools.

\section{Discussion}
\label{sec5}
With the rapid development of artificial intelligence technology, the construction of comprehensive medical diagnostic agents has attracted increasing attention \cite{shen2015emerging}. To the best of our knowledge, Auto-US is currently the first multi-modal ultrasound video diagnostic agent. Our agent achieves SOTA performance in ultrasound disease classification and generates ultrasound reports. 

However, we identify several limitations in existing work. First, we observe that publicly available high-quality ultrasound video datasets are extremely limited. Due to ethical constraints and other reasons, many datasets cannot be publicly released or reused, which hinders the development of advanced diagnostic agents. Additionally, we recognize that, given the complexity of medical diseases, relying solely on ultrasound imaging is insufficient for obtaining accurate diagnostic opinions. Ultrasound images alone cannot serve as the gold standard for disease diagnosis. Another challenge lies in the varying degrees of acceptance of the agent’s diagnostic opinions by different users, which complicates its practical application. 

To build a more powerful ultrasound video diagnostic agent, our future work will involve incorporating larger-scale datasets and more modalities of relevant information.

\section{Method}
\label{sec3}

\subsection{Data Construction}
\label{sec3.1}
\begin{figure}[htbp]
    \centering
    \includegraphics[width=\textwidth]{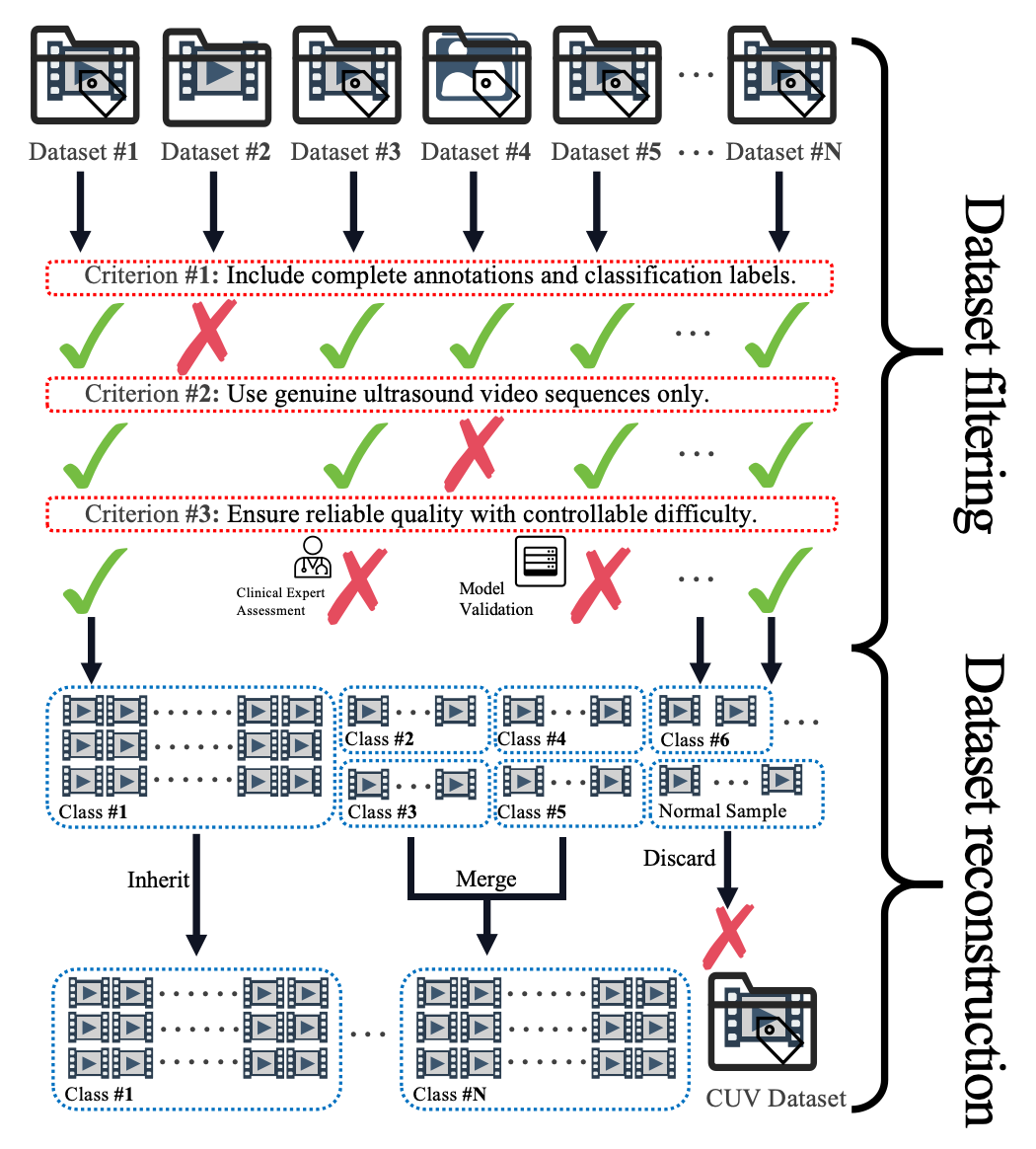} 
    \caption{A flowchart for constructing CUV Dataset.}
    \label{fig2}
\end{figure}

To build a high-quality, large-scale ultrasound video dataset covering multiple anatomical regions and diseases, we thoroughly explored publicly available ultrasound video datasets. The overall pipeline for constructing our dataset is illustrated in Figure~\ref{fig2}.

\subsubsection{Dataset filtering}
\label{sec3.2}
First, to ensure fair comparison across models, we strictly controlled data quality. Our dataset selection was based on the following three criteria: \textbf{1)} The publicly available portion of the dataset must contain complete annotations and classification labels. \textbf{2)} The dataset must consist of genuine ultrasound video sequences, rather than extracted frames or pseudo-video formats. \textbf{3)} The overall data quality must be reliable, with controllable levels of difficulty. To evaluate this aspect, we adopted a two-pronged approach: deep learning-based model validation and clinical expert assessment.

For model validation, we conducted baseline experiments using a standard ResNet architecture. Only datasets that demonstrated adequate classification performance were retained for further analysis. Specifically, we defined a selection criterion that jointly considers classification accuracy and label diversity. Let $Acc$ denote the Top-1 accuracy of the 3D-ResNet model on a given dataset, and $C$ represent the number of distinct categories. A dataset is deemed acceptable if it satisfies the following condition:
\begin{equation}
Acc \geq 1 - \theta \times \log(C)
\end{equation}
where $\theta$ is a scaling factor that controls the trade-off between classification accuracy and category complexity. It is empirically determined based on preliminary experiments; in this study, we set $\theta = 0.4$. This criterion ensures that the dataset is not only learnable by a standard model but also exhibits sufficient categorical diversity to support robust training and fair benchmarking across different algorithms.

For clinical assessment, in parallel, we invited an experienced sonographer to conduct a comprehensive clinical evaluation of the dataset. The assessment focused on several key aspects, including image clarity, diagnostic relevance, and the consistency of annotations. Based on professional expertise and clinical standards, the expert provided feedback on the practical usability of the data. Datasets containing low-quality images, ambiguous diagnostic features, or inconsistent labels were flagged and excluded. This expert-driven assessment complements the quantitative validation and ensures that the final dataset pool reflects real-world diagnostic settings and maintains clinical significance.

\subsubsection{Dataset reconstruction}
\label{sec3.3}
Subsequently, we performed data integration and filtering to construct a fair and broadly representative baseline dataset. Specifically, we merged certain fragmented or overlapping categories. For example, in the gallbladder dataset, some labels were semantically overlapping or clinically ambiguous, which led to misclassification by simple baseline models. To reduce such label noise, we consolidated nine different categories into one unified label.

Furthermore, for diseases with multiple clinical manifestations, such as pneumonia, we retained only key subtypes. Namely, COVID-19 pneumonia and bacterial pneumonia, while excluding normal samples and a small number of viral pneumonia cases that were prone to confusion. It is important to note that the goal of this study is to construct a heterogeneous multi-disease ultrasound video dataset. Therefore, normal samples are treated as a distinct category and should not be excluded from the dataset entirely.

This rigorous filtering process ensures that the final dataset is both clinically meaningful and technically feasible for training and evaluating deep learning models across diverse diagnostic conditions.

\subsection{Classification Model}

To achieve high-precision lesion recognition in ultrasound video, we propose a parallel CNN-Transformer architecture that integrates temporal, spatial, and frequency domain information. Specifically, our network comprises three parallel branches: the \textit{slow path}, the \textit{fast path}, and the \textit{frequency path}.

The \textit{slow path} is designed to accurately extract spatial information from ultrasound videos using a CNN-based architecture. To capture dynamic changes and long-term temporal dependencies in ultrasound sequences, the \textit{fast path} employs a Transformer module that effectively models temporal information.

\begin{figure}[htbp]
    \centering
    \includegraphics[width=\textwidth]{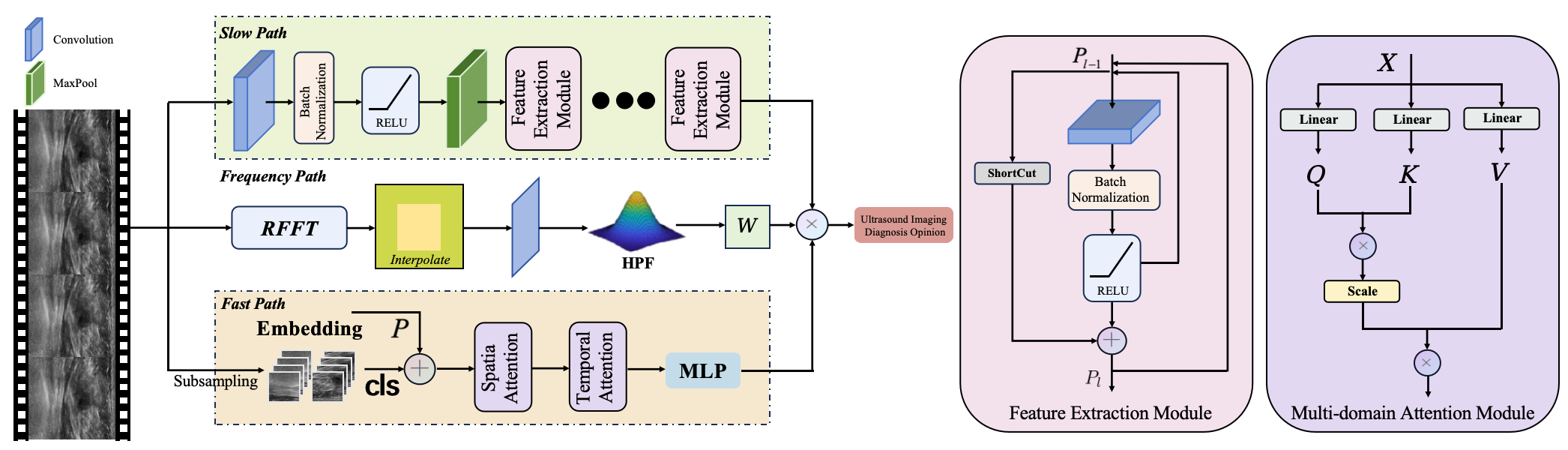} 
    \caption{Architecture diagram of our ultrasound video classification network.}
    \label{fig6}
\end{figure}

In addition to spatial and temporal features, frequency-domain characteristics play a crucial role in medical imaging signals such as ultrasound. Therefore, we introduce the \textit{frequency path}, which leverages a frequency-domain extraction module to perform dynamic fusion of temporal and spatial information with high precision. An overview of the proposed network architecture is illustrated in Figure~\ref{fig6}.

\subsubsection{Slow Path}

To extract comprehensive spatial information from ultrasound videos, we design the \textit{slow path}, which processes the input video using a stack of 3D convolutional layers with residual connections.

In order to clearly express our method, first, let the input video sequence be denoted as $X \in \mathbb{R}^{T \times H \times W \times C}$, where $T$ is the number of frames, $H$ and $W$ represent the spatial dimensions, and $C$ is the number of channels.
The initial feature map $P_0$ defined by the Equation~\ref{eq1}.

\begin{equation}
P_{\text{0}} = \text{MaxPool3D}\left(\text{ReLU}\left(\text{BN}\left(\text{Conv3D}(X)\right)\right)\right),
\label{eq1}
\end{equation}
where, $\text{Conv3D}(\cdot)$ represents 3D convolution operation, $\text{BN}(\cdot)$ represents batch normalization, $\text{ReLU}(\cdot)$ is activation function, and $\text{MaxPool3D}(\cdot)$ is 3D max pooling operation.

Then, the feature map is processed by the feature extraction module, which is defined by the following formula:
\begin{equation}
\begin{aligned}
P_{\ell}^{(1)} &= \text{ReLU}\left( \text{BN}\left( \text{Conv3D}\left(P_{\ell-1}\right) \right) \right),\\
P_{\ell} &= \text{ReLU}\left( \text{BN}\left( \text{Conv3D}\left(P_{\ell}^{(1)}\right) \right) + \text{Shortcut}\left( P_{\ell-1} \right) \right), \quad \text{for } \ell = 1, \dots, N
\end{aligned}
\label{eq:2}
\end{equation}
where  $\text{Shortcut}(\cdot)$ represents the residual connection from the input. And \( N \) is a tunable parameter. In our implementation, we set \( N = 4 \).

By stacking such feature extraction modules, the \textit{slow path} effectively captures the spatial features $F_s = P_N$ from ultrasound videos, demonstrating enhanced sensitivity to small lesions commonly present in medical imaging.

\subsubsection{Fast Path}

To efficiently capture long-term dynamics and fine-grained spatial information in ultrasound videos, we design the \textit{fast path}, which integrates both spatial and temporal attention mechanisms. This branch utilizes a Transformer-based architecture to model sequential dependencies and temporal patterns with high precision.

Firstly, we adopted a down sampling strategy for the input of fast path. Specifically, we define a temporal sampling operator $S(\cdot)$ such that:
\begin{equation}
\begin{aligned}
X_s = S(X; \alpha) = X[::\alpha],
\end{aligned}
\end{equation}
where $\alpha$ is the temporal stride. In our implementation, we choose $\alpha = 5$. 
The subsampled video sequence $X_s \in \mathbb{R}^{T/\alpha \times H \times W \times C}$ is then processed by an embedding layer, which partitions each frame into patches and maps them into a low-dimensional feature space:
\begin{equation}
\begin{aligned}
\{x_1,x_2, \dots ,x_N\} = \mathcal{E}(X_s) = \text{Conv2D}(X_s^t; \text{kernel}=p, \text{stride}=p), t=1,\dots,T/\alpha,
\end{aligned}    
\end{equation}
where $p$ is the patch size. Each frame is divided into $N = \left(\frac{H}{p} \cdot \frac{W}{p}\right)$ patches, resulting in embedded features of shape $x_i \in \mathbb{R}^{1 \times D}$, and $D$ is the embedding dimension.

Next, we introduce a learnable classification token $\mathbf{cls} \in \mathbb{R}^{1 \times D}$ for each frame, and add positional encoding $\mathbf{P} \in \mathbb{R}^{(N+1) \times D}$ to obtain the initial input:
\begin{equation}
\begin{aligned}
z_i = \left[ \mathbf{cls} \, \| \, x_1 \, \| \, \dots \, \| \, x_N \right] + \mathbf{P}.
\end{aligned}  
\end{equation}
To achieve precise temporal modeling of ultrasound videos, we first perform spatial perception to identify important spatial regions, followed by temporal modeling to capture dynamic changes in ultrasound imaging. Specifically, we adopt a spatial attention mechanism, where the features from different frames are first aggregated into $X'_l \in \mathbb{R}^{T \times (N+1) \times D}$. Then, spatial query ($Q$), key ($K$), and value ($V$) vectors are obtained through three independent linear projections, and attention is computed as:

\begin{equation}
\begin{aligned}
Q, K, V &= \text{Linear}(X'_l) \\
Z'_l=\text{Attention}_{\text{spatial}}(Q, K, V) &= \text{Softmax}\left(\frac{QK^\top}{\sqrt{d}}\right)V
\label{eq:7}
\end{aligned}  
\end{equation}

Afterward, the features are re-aggregated along the temporal dimension into $X''_l \in \mathbb{R}^{(N+1) \times T \times D}$, and the same attention operation is applied along the temporal axis. As shown in Equation~\ref{eq:7}, we obtain the temporally modeled representation $Z''_l$. Finally, we enhance the non-linear modeling capacity by applying a Dropout layer and a feed-forward MLP with GELU activation:

\begin{equation}
\begin{aligned}
Z_{l} = Z_l'' + \text{Dropout}(\text{MLP}(\text{LayerNorm}(Z_l''))), \quad \text{for} \quad \ell = 1, \dots, L,
\end{aligned}  
\end{equation}
where, $LayerNorm$ means perform Layer Normalization on the embedding vectors of each token.

After processing all frames, we extract the class token from each frame and apply temporal average pooling to obtain the final aggregated feature in the temporal domain:

\begin{equation}
\begin{aligned}
F_f = \frac{1}{T} \sum_{t=1}^{T} Z_L[:, t, 0, :].
\end{aligned}  
\end{equation}

This design enables the fast path to rapidly capture and analyze temporal variations in ultrasound videos, complementing the spatial feature extraction performed by the slow path. By modeling temporal dynamics, our network effectively identifies changes associated with ultrasound lesions, which is critical for accurate and dynamic lesion diagnosis.

\subsubsection{Frequency Path}

To jointly capture spatial and temporal information from ultrasound videos, we introduce the \textit{frequency path}, which enables dynamic spatiotemporal feature integration. This path enhances classification performance by leveraging spectral domain representations of video frames.

Each input frame is first transformed using a 2D Fast Fourier Transform to obtain its complex spectrum:
\begin{equation}
\mathcal{F}(X) = \mathrm{RFFT2}(X), \quad \mathcal{F}(X) \in \mathbb{C}^{H \times (W/2+1)},
\end{equation}
where $\mathcal{F}(X)$ denotes the complex-valued frequency spectrum obtained via the 2D real-input FFT of the real-valued input frame $X \in \mathbb{R}^{H \times W}$.  
The magnitude spectrum $M$ is then extracted and upsampled to match the original spatial resolution:
\begin{equation}
M = \mathrm{Interpolate}(|\mathcal{F}(X)|, \text{size}=(H, W)).
\end{equation}
Note that $|\mathcal{F}(X)|$ denotes the element-wise complex magnitude of the frequency spectrum.

Subsequently, a 2D convolutional layer is applied to extract spectral features $F_m$, followed by a high-pass filter to enhance edge and detail responses:
\begin{equation}
\begin{aligned}
F_m = \text{Conv2D}(M), \quad F_h = \text{HPF}(F_m),
\end{aligned}  
\end{equation}
where the high-pass filter $\text{HPF}$ uses a fixed Laplacian kernel:
\begin{equation}
\begin{aligned}
K = 
\begin{bmatrix}
0 & -1 & 0 \\
-1 & 4 & -1 \\
0 & -1 & 0
\end{bmatrix}.
\end{aligned}  
\end{equation}

This convolution is applied independently across channels to preserve per-channel frequency responses.

We then perform max pooling followed by adaptive average pooling on the high-frequency features to obtain compact representations. Resulting in a feature map $F_g \in \mathbb{R}^{T \times d}$. The features are reshaped into video format $[T, d]$ and averaged across time to produce the frequency-guided representation:
\begin{equation}
\begin{aligned}
F_{\text{freq}} = \frac{1}{T} \sum_{t=1}^{T} F_g^{(t)}.
\end{aligned}  
\end{equation}

To guide the fusion of temporal features from both the slow and fast paths, a lightweight MLP is used to generate fusion weights. The output is normalized using a softmax function:
\begin{equation}
\begin{aligned}
[\alpha_s, \alpha_f] = \text{Softmax}(W_f  \cdot F_{\text{freq}}), \quad \text{with} \quad \alpha_s + \alpha_f = 1,
\end{aligned}  
\end{equation}
where $W_f$ is the learnable weight matrix of the MLP layer. The frequency path serves as a modulation module, providing prior guidance for dynamic fusion of spatial and temporal features.

Given the feature outputs of the slow and fast paths:$F_s \in \mathbb{R}^{B \times D}$, $F_f \in \mathbb{R}^{B \times D}$, the final fused feature is obtained through weighted summation:
\begin{equation}
\begin{aligned}
F_{\text{fused}} = \alpha_s \cdot F_s + \alpha_f \cdot F_f.
\end{aligned}  
\end{equation}

This fusion strategy adaptively balances the contributions of spatial and temporal features based on the frequency characteristics of the input.

Finally, the fused representation is passed to a linear classifier for prediction:
\begin{equation}
\begin{aligned}
\hat{y} = \text{Softmax}(W_c \cdot F_{\text{fused}} + b),
\end{aligned}  
\end{equation}
where $W_c \in \mathbb{R}^{C \times D}$ is a learnable classification head and $C=5$ denotes the number of categories.

The overall model integrates features from the three parallel paths through a weighted fusion mechanism. The fusion weights are generated by frequency paths, which takes the concatenated features from the slow and fast paths as input. This design enables the model to adaptively balance the contributions of different branches, leveraging the complementary information from temporal and frequency domains to achieve accurate ultrasound video classification.

\subsection{Integration of Large Language Models}

\begin{figure}[htbp]
    \centering
    \includegraphics[width=\textwidth]{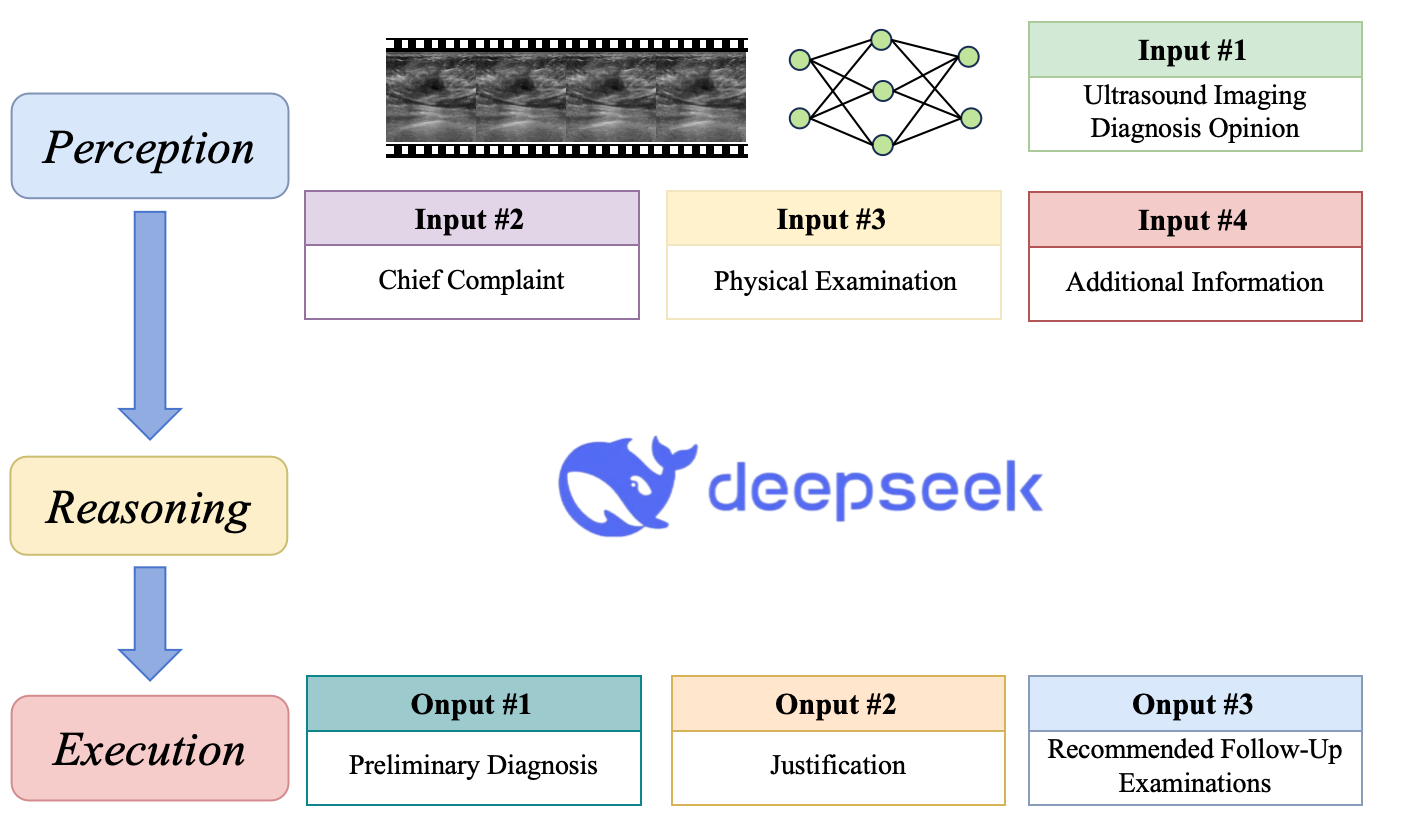} 
    \caption{Workflow diagram of Auto-US Agent.}
    \label{fig5}
\end{figure}

Our study proposes an automated ultrasound diagnostic agent system that integrates deep learning with LLMs, aiming to assist clinicians in more accurate and efficient ultrasound video classification and lesion assessment. The system adheres to a modern agent construction paradigm and encompasses three core stages: video perception, clinical reasoning, and diagnostic decision-making. It achieves a closed-loop diagnostic workflow from image recognition to clinical diagnostic generation. The detailed pipeline is shown in Figure \ref{fig5}.

\subsubsection{Construction Workflow}

To develop an effective and user-friendly diagnostic agent, we propose a three-stage architecture comprising \textit{perception}, \textit{reasoning}, and \textit{execution}. In the perception stage, a high-precision ultrasound video classification network is employed to obtain accurate disease categories.

For the reasoning stage, we leverage the ultrasound lesion classification network introduced in the previous section to produce structured diagnostic labels. These labels, combined with physical examination data, are formatted into a structured prompt and fed into the large language model \textit{DeepSeek-R1-7B} \cite{guo2025deepseek}.

As the cognitive core of the agent, DeepSeek-R1-7B exhibits strong capabilities in logical reasoning and medical language comprehension \cite{sandmann2025benchmark}. In the execution stage, the agent generates standardized diagnostic suggestions based on the content of the prompt. The final output includes a diagnostic report, risk stratification, and follow-up recommendations. It also highlights regions of uncertainty based on model confidence and assigns diagnostic categories in accordance with international clinical guidelines to assist in medical decision-making.

\subsubsection{Prompt Template Design}

To ensure the reliability and standardization of the diagnostic suggestions generated by the agent, we design prompt templates grounded in internationally recognized clinical guidelines. The guidelines incorporated into the prompts primarily include the following standards:

\begin{itemize}
    \item \textbf{Pneumonia}: Based on the IDSA/ATS Guidelines for the Diagnosis and Treatment of Community-Acquired Pneumonia in Adults (2019).
    \item \textbf{Gallbladder Diseases}: Based on the American College of Gastroenterology (ACG) Guidelines for the Management of Gallbladder Diseases (2022).
    \item \textbf{Breast Diseases}: Based on the Breast Imaging Reporting and Data System (BI-RADS), 5th Edition.
\end{itemize}

To facilitate ease of use while ensuring accurate and well-structured outputs, we construct a task-structured universal prompt template that flexibly integrates classification results and clinical input information:

\begin{quote}
\small
You are a senior ultrasound imaging diagnostic expert. Based on the information below, determine the patient's possible condition and provide diagnostic recommendations.

\begin{itemize}
  \item \textbf{Ultrasound Imaging Diagnosis Opinion:} \texttt{<Model\_Result>}
  \item \textbf{Chief Complaint:} \texttt{<Chief\_Complaint>}
  \item \textbf{Physical Examination:} \texttt{<Physical\_Exam>}
  \item \textbf{Additional Information:} \texttt{<Additional\_Info>}
\end{itemize}

Please reason according to international diagnostic guidelines and generate a medically standardized recommendation using professional terminology. The output format should be:

\begin{itemize}
  \item \textbf{Preliminary Diagnosis:}
  \item \textbf{Justification:}
  \item \textbf{Recommended Follow-Up Examinations:}
\end{itemize}

\end{quote}

\subsubsection{Evaluation Protocol}

To comprehensively assess the practicality of the diagnostic recommendations generated by the LLM, we employ both qualitative and quantitative evaluation strategies.

For \textit{qualitative analysis}, each output sample is independently evaluated by 2 ultrasound imaging specialists and 3 non-expert volunteers using a 5-point Likert scale, where: \textbf{1} indicates a useless diagnostic suggestion; \textbf{2}, only relevant keywords are correct; \textbf{3}, marginally acceptable; \textbf{4}, useful for diagnostic assistance; and \textbf{5}, comparable to a professional clinician’s opinion.

For \textit{quantitative analysis}, we adopt the Metric for Evaluation of Translation with Explicit Ordering (METEOR) score \cite{banerjee2005meteor}. It is a widely used language evaluation metric that considers precision, recall, stemming, synonymy, and word order alignment, to measure the semantic similarity between LLM-generated suggestions and reference diagnoses. Unlike surface-level metrics such as BLEU, METEOR is designed to better correlate with human judgment by capturing both lexical and semantic matching \cite{chauhan2023comprehensive}.

To provide a more comprehensive evaluation of the agent, we designed a weighted scoring framework:

\begin{equation}
\text{Final Score} = 0.2 \times S_{\text{amateur}} + 0.6 \times S_{\text{expert}} + 0.2 \times 5 \times M,
\end{equation}
where $S_{\text{amateur}}$ and $S_{\text{expert}}$ represent the average scores from amateurs and experts, respectively, and $M$ denotes the normalized METEOR score in the range $[0,1]$.

This hybrid evaluation framework effectively assesses the real-world utility of Auto-US in assisting ultrasound diagnostics.

The framework highlights how Auto-US, leveraging the synergy between deep learning and large language models, can deliver guideline-compliant, interpretable, and high-quality diagnostic recommendations in ultrasound imaging.

\section*{Data availability}
All datasets utilized in this study are publicly available as open-access resources. A comprehensive guide for accessing and using the data is provided at \url{https://github.com/Bean-Young/Auto-US}. The code and clinical data used in this study is also available at the same link.

\section*{Ethical statement}
For the comprehensive ultrasound dataset, all data used in this study were obtained from open-access datasets and were fully anonymized prior to use. These datasets are secondary in nature and contain no identifiable personal information. All data were securely managed and de-identified to protect the privacy and rights of individuals.

This study was approved by the Biomedical Ethics Committee of Anhui University (Approval No.BECAHU-2024-023) and was conducted in accordance with the Declaration of Helsinki (1964) and its later amendments or comparable ethical standards. As the study involved no direct interaction with individuals and no identifiable personal data, the requirement for informed consent was waived by the Institutional Review Board.

For clinical data cases, this study was reviewed and approved by the Ethics Review Committee of Shanghai Fengxian District Central Hospital. All procedures involving human participants were performed in accordance with the ethical standards of the 1964 Declaration of Helsinki and its later amendments. Informed consent was waived by the ethics committee due to the retrospective nature of the study, which involved analysis of anonymized clinical data (Approval No.SL2024-KY-29-01).

\section*{Author Contributions}
Y.Y. conceptualized the study, developed the methodology, performed formal analysis, wrote the original draft, reviewed and edited the manuscript, and was responsible for coding and visualization. Y.G. assisted in writing the original draft and was responsible for visualization. W.C. contributed to validation, data curation, coding, and writing the original draft. Q.R. provided resources and contributed to data curation. S.W. conducted the investigation and was responsible for data curation. X.D., Z.J., and Y.D. were involved in supervision, project administration, and funding acquisition. All authors have read and agreed to the published version of the manuscript.

\bibliography{ref}

\end{document}